\documentclass[runningheads]{llncs}
\usepackage{hyperref}
\usepackage{graphicx}
\usepackage{amsmath,amssymb} % define this before the line numbering.
\usepackage{color}
\usepackage{subfigure} 
\usepackage{algorithm}
\usepackage{algorithmic}

\graphicspath{{./figs/}}

\usepackage{mathtools,lipsum}

\begin{document}
% \linenumbers
\pagestyle{headings}
\mainmatter
\def\ECCV18SubNumber{2910}  % Insert your submission number here

\title{Deep Randomized Ensembles for Metric Learning} % Replace with your title
\titlerunning{Deep Randomized Ensembles for Metric Learning}

\author{Hong Xuan$^1$\href{https://orcid.org/0000-0002-4951-3363}, Richard Souvenir$^2$, Robert Pless$^1$}
\authorrunning{Hong Xuan, Richard Souvenir, Robert Pless}

\institute{$^1$Department of Computer Science,\\
	George Washington University\\
	\email{ \{xuanhong,pless\}@gwu.edu} \\
$^2$Department of Computer and Information Sciences,\\
Temple University\\
\email{souvenir@temple.edu}
}

\maketitle

\begin{abstract}
Learning embedding functions, which map semantically related inputs to nearby locations in a feature space supports a variety of classification and information retrieval tasks.  In this work, we propose a novel, generalizable and fast method to define a family of embedding functions that can be used as an ensemble to give improved results.  Each embedding function is learned by randomly bagging the training labels into small subsets.  We show experimentally that these embedding ensembles create effective embedding functions. The ensemble output defines a metric space that improves state of the art performance for image retrieval on CUB-200-2011, Cars-196, In-Shop Clothes Retrieval and VehicleID. Code is available at:  \textcolor{magenta}{https://github.com/littleredxh/DREML}

\end{abstract} 

%%%%%%%%%%%%%%%%%%%%%%%%%%%%%%%%%%%%%%%%%%%%%%%%%%%%%%%%%%%%%%%%%%%%%%%%%%%%%%%%%%%%%%%%%%%%%%%%%%%%%%%%%%%%%%%%%%%%%%%%%%%%%%%%%%%%%%%%%%%%%%%%%%%%%%

\section{Introduction}
\label{01Intro}
% Embedding is an important problem.
Image embeddings are commonly optimized to map semantically similar inputs to nearby locations in feature space. Thereafter, tasks such as classification and image retrieval can be recast as simple operations, such as neighborhood lookups, in the learned feature space.  This approach has been applied across many problems.

Deep learning approaches to embedding are trained with data from many classes and optimize loss functions based on pairs of images from the same or different classes~\cite{chopra2005learning,hadsell2006dimensionality}, triplets of images where inputs from the same class are forced to be closer than inputs from different classes~\cite{schroff2015facenet}, or functions of large collections of images~\cite{law2013quadruplet,chen2017beyond,ustinova2016learning,HermansBeyer2017Arxiv}.

For many of these optimization functions, embedding the input images into a high 
dimensional space leads to poor performance due to over-fitting. Some recent work~\cite{BIER} suggests an approach to high-dimensional embedding with an ensemble approach, learning to map images to a collection of independent output spaces, using boosting to re-weight input examples to make each output space independent.

\begin{figure}[t]
    \centering
    \includegraphics[height=3.5in]{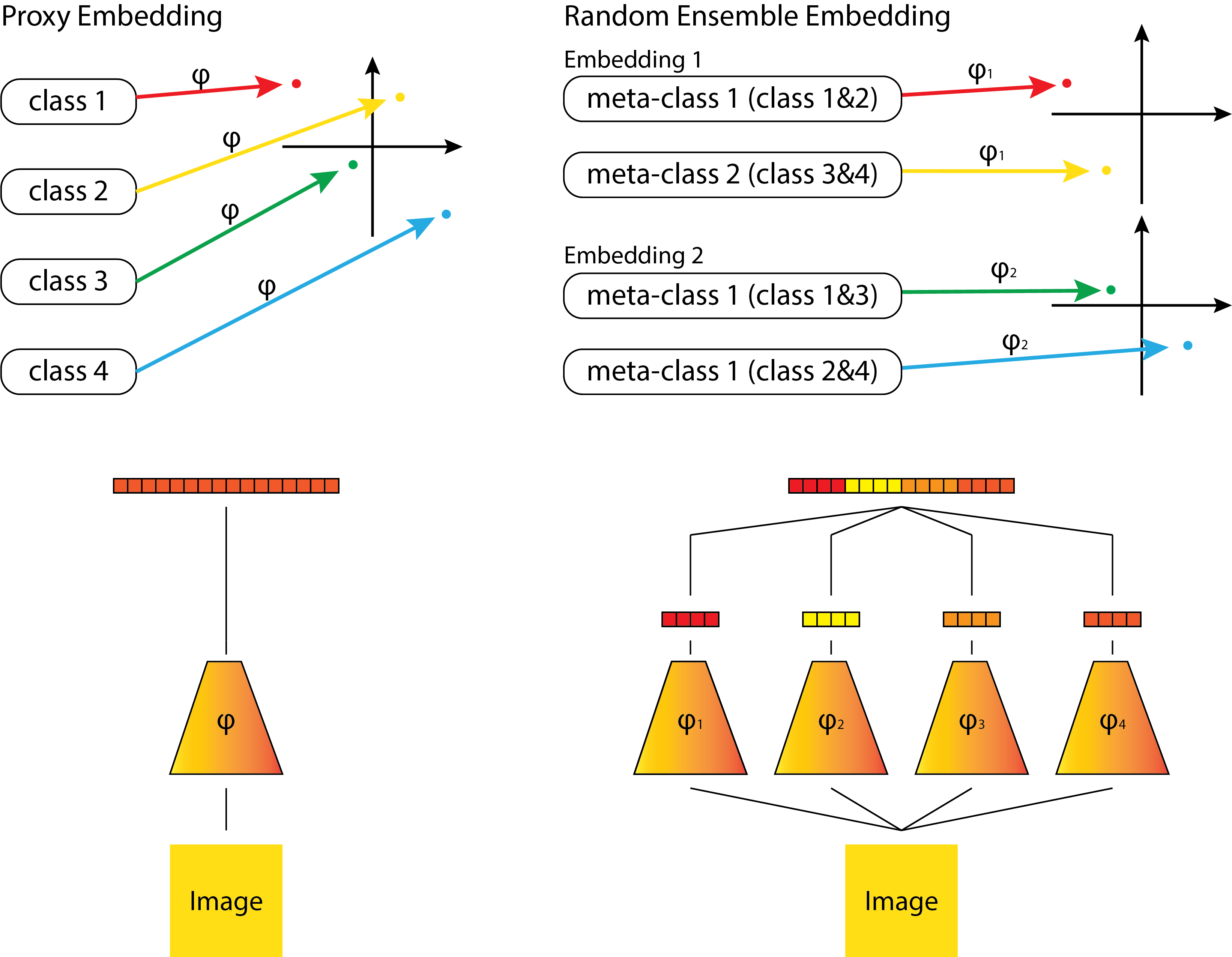}
    \caption{Standard deep embedding approaches (left) train a network to map all images from a class to nearby locations in an output space. Our approach (right) learns an ensemble of mappings.  Each model of the ensemble learns a mapping that groups small subsets of classes. Images are mapped by each model in the ensemble, and the output coordinates are concatenated.}
    \label{fig:teaser}
\end{figure}

\begin{figure}[t]
\vskip 0.2in
\begin{center}
{\includegraphics[width=\columnwidth]{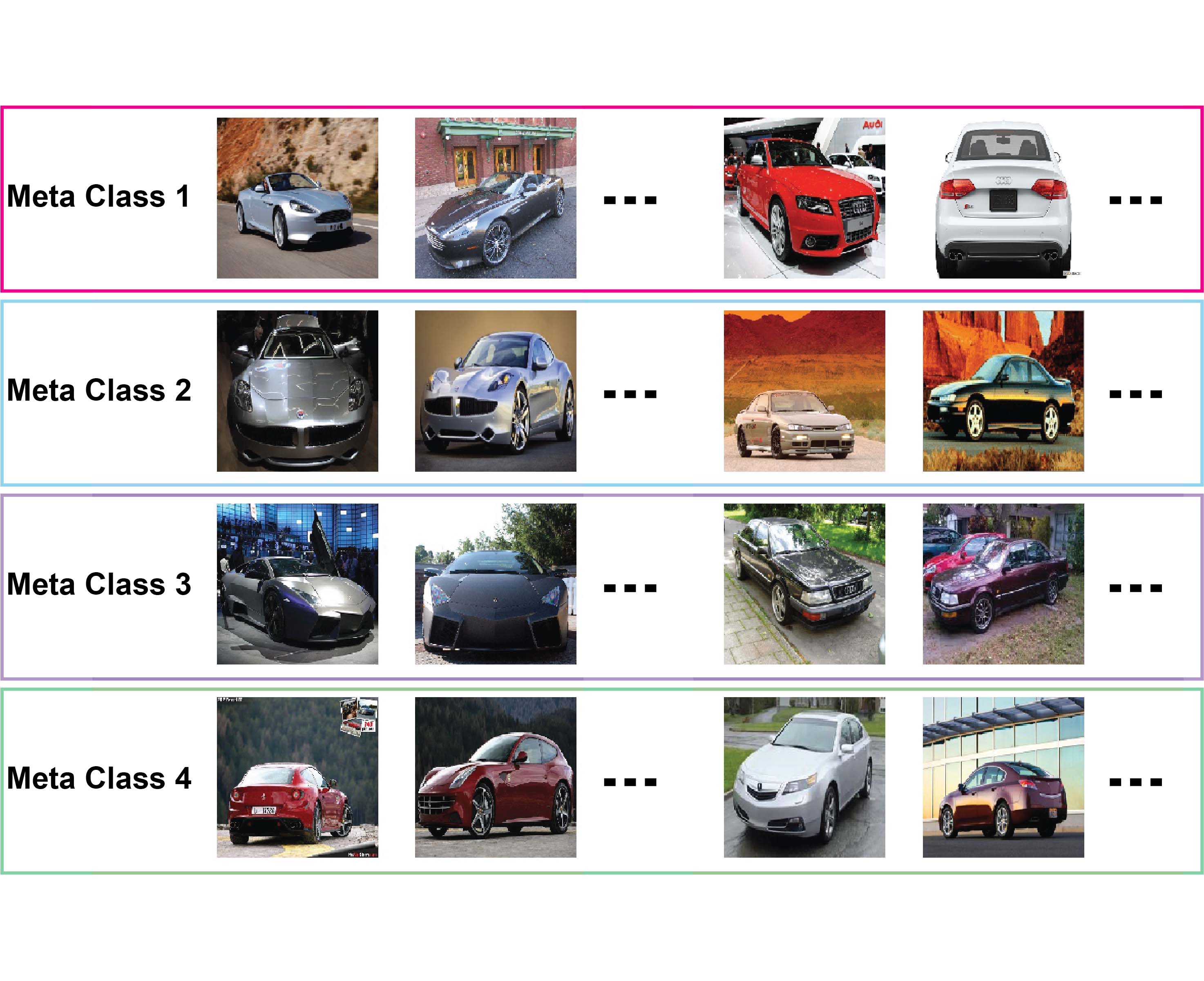}}
\caption{Meta-classes for the CAR196 dataset. It may be counter-intuitive that it is helpful to group specific models of Porches and Audis into one class and learn embeddings where those classes are mapped to the same location, but this approach makes it easy to define many different but related embedding problems that become an effective ensemble.}
\label{metaClassFigure}
\end{center}
\vskip -0.2in
\end{figure}

We propose a different approach to learning a robust, high-dimensional embedding space.  Instead of re-weighting input examples to create independent output embeddings, we propose to group class labels.  Figure~\ref{fig:teaser} illustrates the idea.  We learn a collection of embeddings all trained with the same input data, but differing in the label assigned to the data points.  We group classes into meta-classes (each containing a few classes), and learn embeddings with inputs labelled by their meta-class. We show a visual example of the meta-classes in Figure~\ref{metaClassFigure} based on data from the CARS196 dataset.  The first meta-class groups together images from particular models of Porches and Audi, so the first embedding will seek to map all these images to the same location.

For each grouping of classes into meta-classes, an embedding is learned that embeds all elements of the same meta-class similarly.  We train many such embeddings, with different random groupings of classes into meta-classes. The final embedding is the concatenation of the coordinates of each low-dimensional embedding. 

This approach fits to many choices of embedding architectures.  We show experimental results using the ResNet-18~\cite{resnet} and Inception V1~\cite{Szegedy_2016_CVPR} architectures, and demonstrate that our ensemble method improves upon the state of the art across a number of problem domains. Our contributions are as follows.  
\begin{itemize}
    \item We introduce the idea of randomly grouping labels as an approach to making a large family of related embedding models that can be used as an ensemble.
    \item We illustrate the effect of different parameter choices relating to the embedding size and number of embeddings within the ensemble.
    \item We demonstrate improvement over the state of the art for retrieval tasks for the CUB-200-2011~\cite{CUB200}, Cars-196~\cite{CAR196}, In-Shop Clothes Retrieval~\cite{ICR} and VehicleID~\cite{VID} datasets.
\end{itemize}

%%%%%%%%%%%%%%%%%%%%%%%%%%%%%%%%%%%%%%%%%%%%%%%%%%%%%%%%%%%%%%%%%%%%%%%%%%%%%%%%%%%%%%%%%%%%%%%%%%%%%%%%%%%%%%%%%%%%%%%%%%%%%%%%%%%%%%%%%%%%%%%%%%%%%%
\section{Related Work}
\label{02Related}
\subsection{Image Embeddings}

Image embedding falls under the umbrella of distance metric learning. There has been quite a bit of work in this area from both the machine learning and computer vision perspectives. Here, we focus on recent methods which employ convolutional neural networks for image embedding.

There are many ways to learn embedding functions.  Triplet loss (e.g.~\cite{schroff2015facenet}) defines a loss function based on triplets of images (two from the same class, and one from a different class), and penalizes the network if it does not map the same class inputs to be closer than the different classes.  Because training is often performed in batches, it is natural to consider loss functions that optimize the embedded location of all images in the batch, either by considering all triplets (defined by the batch) simultaneously or by penalizing the histograms of distance between same-class and different class images~\cite{HermansBeyer2017Arxiv,lifted,ustinova2016learning}.

No Fuss Embeddings~\cite{Proxy} shows that using the output layer of classification networks provide very useful embedding functions for one-shot learning and image retrieval tasks.  This has the advantage of faster convergence (because each input image has a specific label, and the loss function does not depend on where other inputs are mapped), which removes some challenges in hard-example mining that plague some triplet-loss approaches.  While triplet-loss approaches can be designed to mitigate these challenges~\cite{HermansBeyer2017Arxiv}, we choose to use~\cite{Proxy} as our embedding approach in our experiments primarily for its speed.

\subsection{Ensemble CNNs}
Ensemble algorithms have been more widely used for classification problems. One example applies a variation of the boosting model that adds extra network layers trained on examples for which a smaller network fails~\cite{DBLP:journals/corr/YuanYZ16}.  Other approaches to create diversity in the ensemble is to train a collection of networks of different architectures to solve the same problem and combine the results~\cite{guo2015deep}.

To the best of our knowledge, the only work that creates an ensemble for embedding is BIER~\cite{BIER}. This follows a boosting model to incrementally create an ensemble embedding by re-weighting examples so that subsequent embeddings are driven to correct errors in earlier embeddings. Compared to this approach, our method is not sequential, and therefore trivially parallelized. Additionally, as we demonstrate in Section~\ref{Experiment}, our method outperforms BIER on many benchmark datasets. 

%%%%%%%%%%%%%%%%%%%%%%%%%%%%%%%%%%%%%%%%%%%%%%%%%%%%%%%%%%%%%%%%%%%%%%%%%%%%%%%%%%%%%%%%%%%%%%%%%%%%%%%%%%%%%%%%%%%%%%%%%%%%%%%%%%%%%%%%%%%%%%%%%%%%%%
\section{Training Randomized Ensemble Embedding}

Our training approach is to create a collection of related models and learn an embedding for each one. To create one member of our ensemble, model $i$, we partition the set of class labels from the training set $Y$ into a set of meta-classes $M_i$, where the number of meta-classes is a parameter $D$, and each meta-class is roughly the same size. All model in the ensemble are computed the same way, with the only difference being that the mapping, $\phi_i$, based on meta-classes, $M_i$, come from a different random partitions of $Y$.

%, so $M_i = \{M_{i1}, M_{i2}, M_{i3}, \ldots M_{iD} \}$, and $|M_{i:}| \approx \frac{|Y|}{D}$.

We define the number of embeddings as $L$. To compute the final embedding for a new input $x$, we concatenate the output of each embedding to get a final output vector, $\Phi = \langle \phi_1(x), \phi_2(x), \ldots \phi_L(x)\rangle$.  For one-shot learning or image retrieval tasks, this function $\Phi$ takes the place of standard embedding functions.  Overall this approach has a collection of parameters and choices, with the two most prominent being: 
\begin{enumerate}
    \item $D$, the number of meta-classes into which the class label set $Y$ is partitioned, and
    \item $L$, the number of embeddings functions included in the ensemble.
\end{enumerate}
There are a collection of choices that relate to learning $\phi_i$ based on the embedding problem defined by the class partition $M_i$.  If $\phi_i$ is represented as a deep neural network, we consider the following questions: 
\begin{enumerate}
    \item What is output embedding dimension of $\phi_i$?
    \item What is the network architecture that represents the function $\phi_i$?
    \item What is the loss function used to train $\phi_i$?
\end{enumerate}

We chose to experiment with ResNet-18 and Inception V1 architectures and follow the no-fuss-embedding approach~\cite{Proxy} with an output dimension equal to the number of meta-classes, $D$ (except where noted). Given these choices, Section~\ref{Experiment} characterizes performance as a function of the number of meta-classes (and therefore the size of each meta-class) and the size of the ensemble used.

% \subsection{Implementation Details}
% \label{Implementation}

All tests are run on the PyTorch platform~\cite{pytorch}.  For our experiments, we use the ResNet18 and Inception V1 implementations, which are pretrained on ILSVRC 2012-CLS data~\cite{ILSVRC15}. The input images are re-sized to 256 by 256 pixels.  We adopt a standard data augmentation scheme (random horizontal flip and random crops padded by 10 pixels on each side). For pre-processing, we normalize the images using the channel means and standard deviations. All networks are trained using stochastic gradient descent (SGD). On all datasets we train using a batch size of 128 for 12 epochs. The initial learning rate is set to 0.01 and divided by 10 every 4 epochs.

%%%%%%%%%%%%%%%%%%%%%%%%%%%%%%%%%%%%%%%%%%%%%%%%%%%%%%%%%%%%%%%%%%%%%%%%%%%%%%%%%%%%%%%%%%%%%%%%%%%%
\section{Experimental Evaluation}
\label{Experiment}

We compare our method, Deep Randomized Ensembles for Metric Learning (DREML) with 7 state-of-art methods (using published results where available): Triplet Learning with semi-hard negative mining~\cite{triplet}, N-Pairs deep metric loss~\cite{Npair}, Proxy-based method~\cite{Proxy}, Hard-Aware Deeply Cascaded Embedding (HDC)~\cite{HDC}, Boosting Independent Embeddings Robustly (BIER)~\cite{BIER}, the FashionNet benchmark~\cite{ICR} and Group Sensitive Triplet Sampling (GS-TRS)~\cite{DBLP:journals/corr/BaiGLWHD17}.

\subsection{Parameter Selection}
\label{Effect}
\begin{figure}
\begin{center}
{\includegraphics[width=0.48\columnwidth]{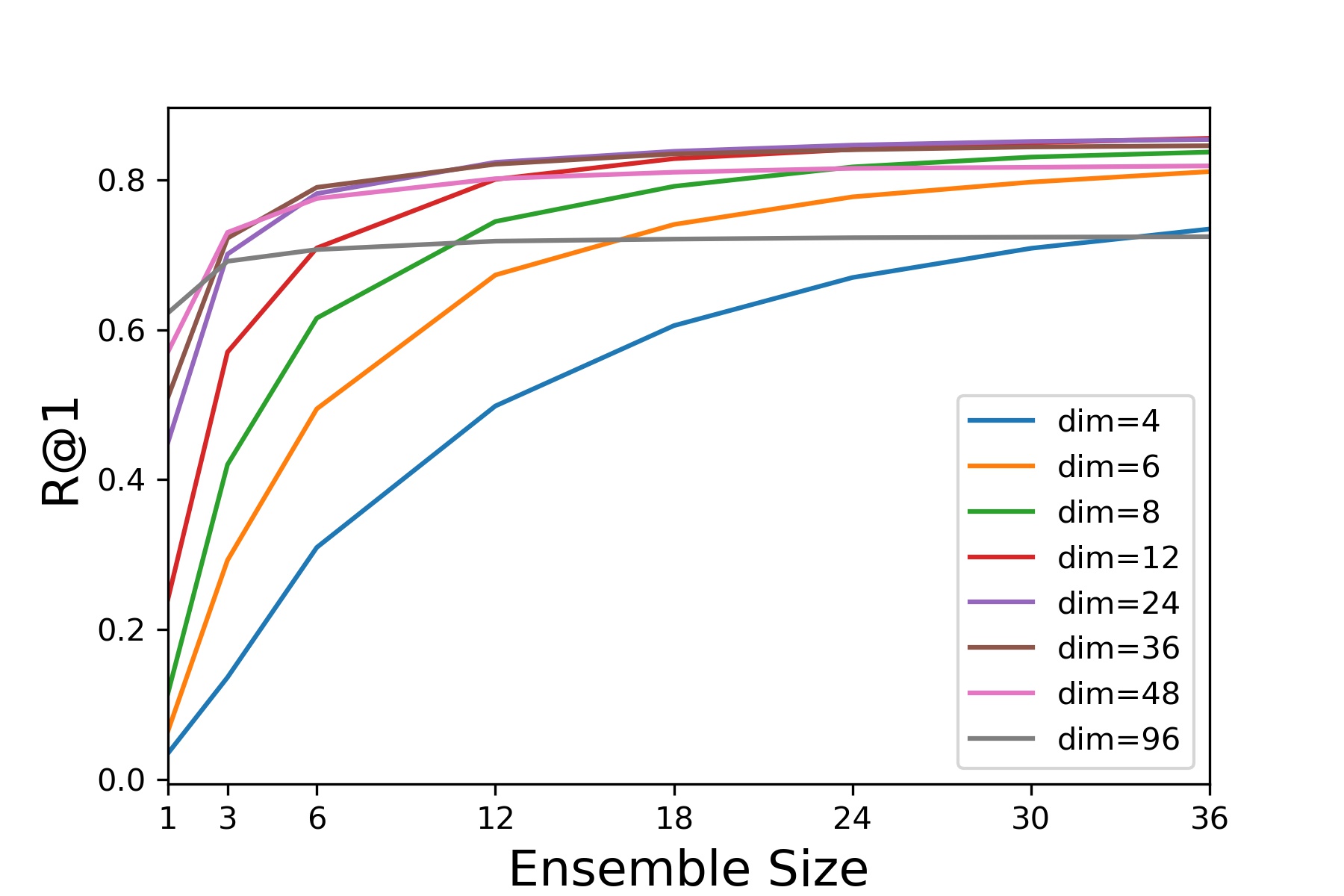}}
{\includegraphics[width=0.48\columnwidth]{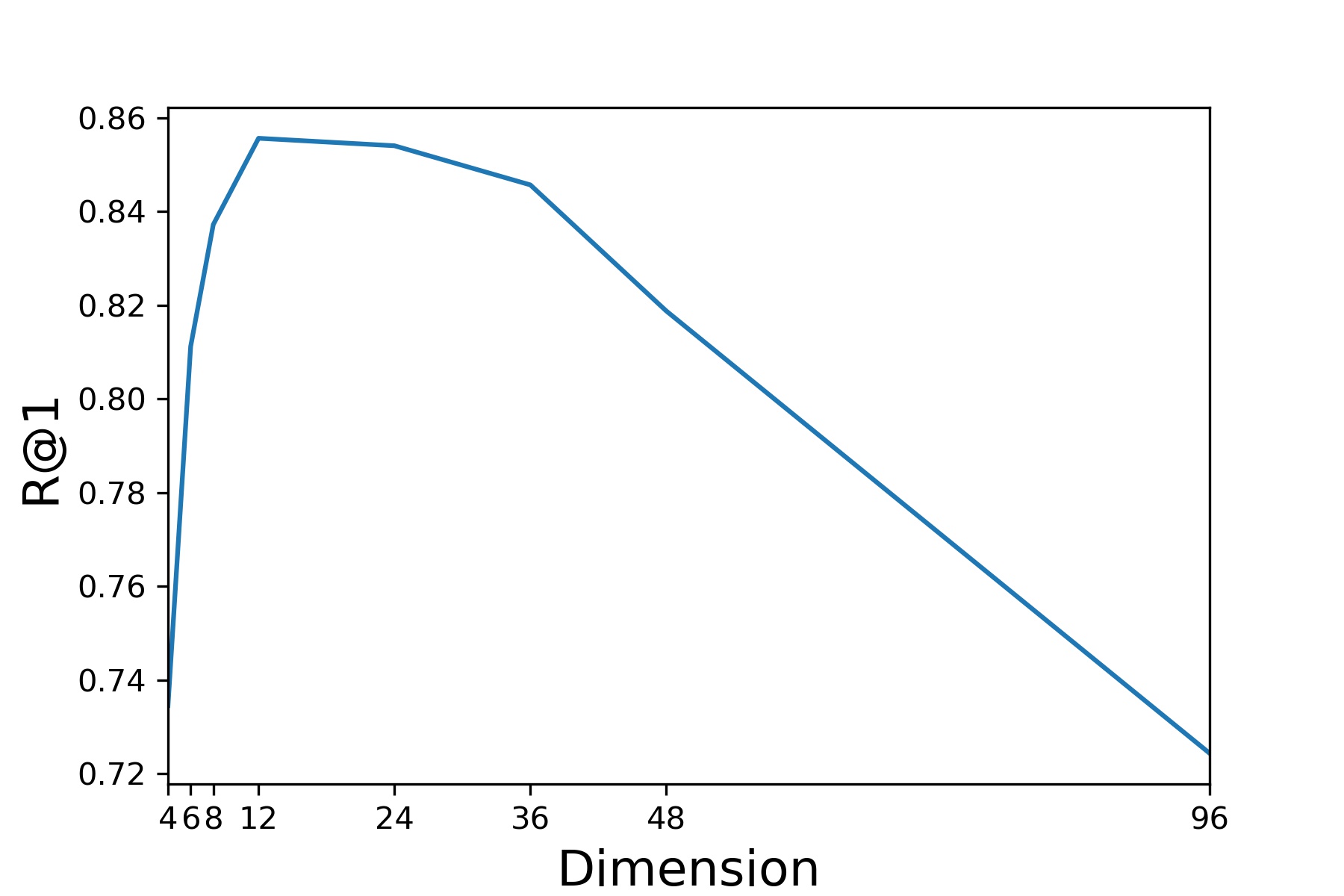}}
\caption{Left: Recall@1 accuracy on the CAR196 dataset with various $D$ (different lines) and $L$ (on the x-axis). Right: Recall@1 accuracy of the largest ensemble models with various $D$. Extreme cases have poorer performance because the individual models must deal with either many meta-classes per class (small $D$) or a lack of diversity (large $D$).}
\label{D_Figure}
\end{center}
\end{figure}

Figure~\ref{D_Figure} shows the performance tradeoffs for different choices of the number of embeddings to include in the ensemble (our parameter $L$), and the number of meta-classes (our parameter $D$), for the CAR196 dataset. The left graph shows a dramatic improvement as the ensemble size grows while it is small, and a clear asymptotic behavior beyond which adding new embeddings does not help. The right graph shows that the performance also depends on the size of the meta-class. When D is small, the number of classes per meta-class is large making a harder embedding problem; when D is large, the number of classes per meta-class is small leading to less diversity in the ensemble. 

\begin{figure}
\begin{center}
{\includegraphics[width=0.32\columnwidth]{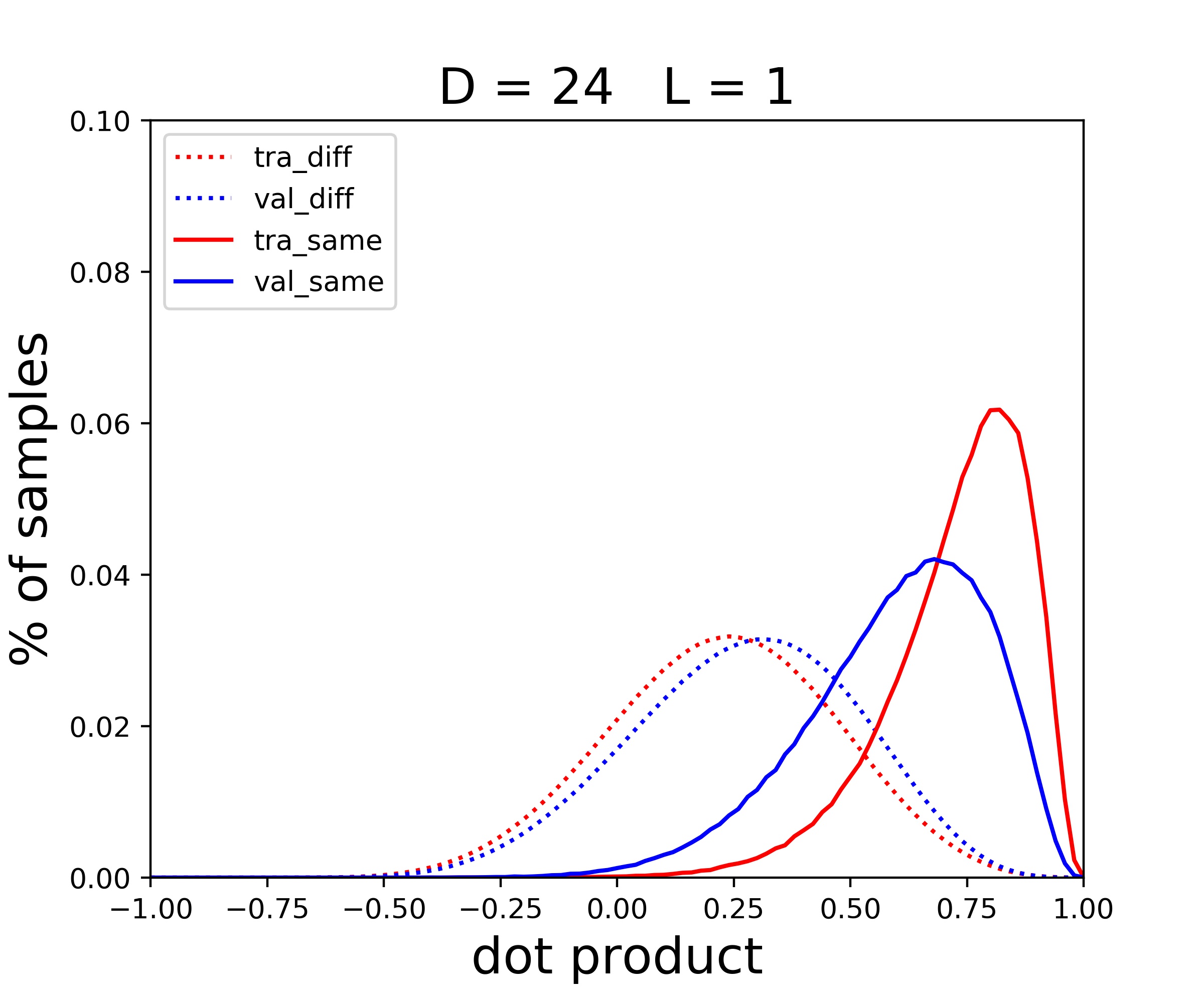}}
{\includegraphics[width=0.32\columnwidth]{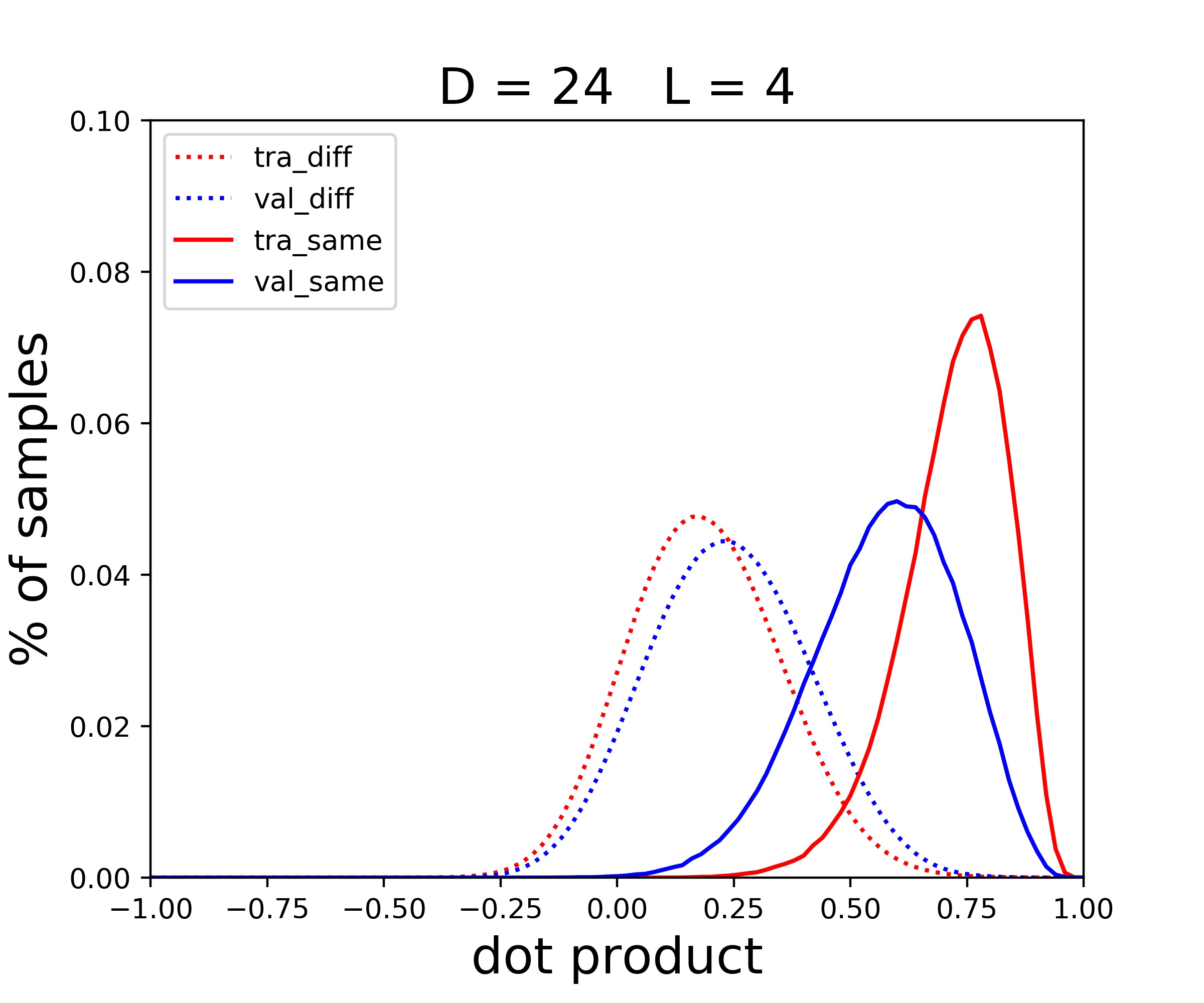}}
{\includegraphics[width=0.32\columnwidth]{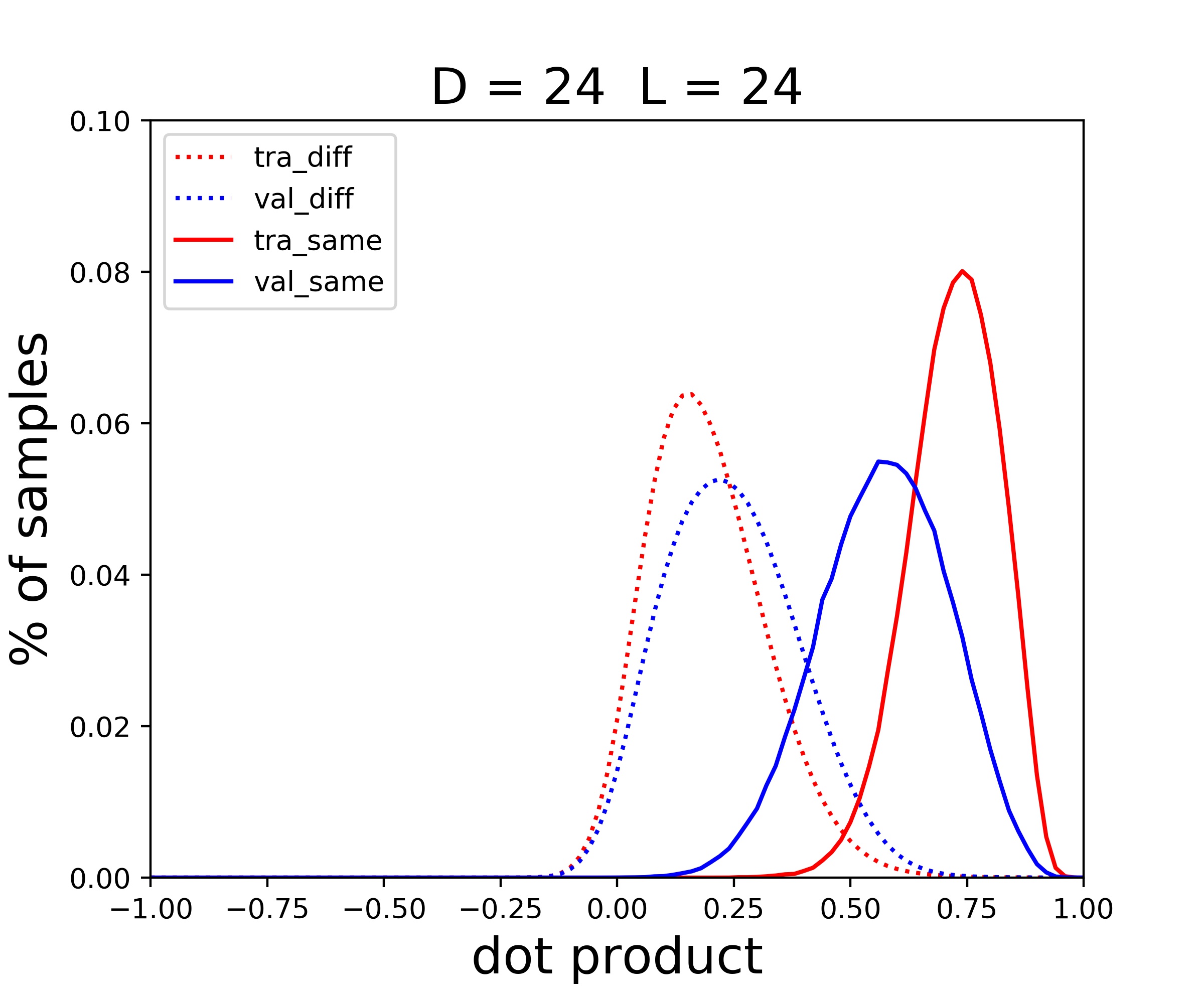}}
\caption{Pair-wise dot product distribution of image feature vectors in the same classes (solid) and in different classes (dashes), for images from the training classes (red) and validation classes (blue).  Shown are the distributions for 1, 4, 24 network ensembles, all with an output dimension of 24.
As the number of networks grow, the distributions for same and different categories separate.}
\label{L_Figure}
\end{center}
\end{figure}

Figure~\ref{L_Figure} explores the effect of increasing ensemble size for a fixed meta-class size.  We see that the distribution of dot-products between embeddings of objects in the same class (solid) and different classes (dashed) becomes more separated for the CAR196 dataset, for both the training and validation datasets.  Additionally, the number of pairs from different classes that have a large dot-product (for example, greater than 0.75) decreases.  This is consistent with the observed improvement in the recall performance. 

For the remaining experiments, we employ multiple DREML models, denoted as DREML (\{I,R\}, D, L) where the tuple indicates the architecture (I)nceptionV1 or (R)esNet18 and values for $D$ and $L$. 

\begin{table}
\caption{Retrieval and Clustering Performance on the CUB200 and CAR196 dataset.}
\begin{center}
\begin{tabular}{|l|ccccc|ccccc|}
\hline
& \multicolumn{5}{|c|}{CUB200} & \multicolumn{5}{|c|}{CAR196}\\
\hline
Method & R@1 & R@2 & R@4 & R@8 &NMI & R@1 & R@2 & R@4 & R@8 &NMI\\
\hline
TRIPLET    
& 42.6 & 55.0 & 66.4 & 77.2 & 55.4 & 51.5 & 63.8 & 73.5 & 81.4 & 53.4\\
N-PAIRS
& 51.0 & 63.3 & 74.3 & 83.2 & 60.4 & 71.1 & 79.7 & 86.5 & 91.6 & 64.0\\
PROXY
& 49.2 & 61.9 & 67.9 & 72.4 & 59.5 & 73.2 & 82.4 & 86.4 & 88.7 & 64.9\\
HDC
& 53.6 & 65.7 & 77.0 & 85.6  & $-$ & 73.7 & 83.2 & 89.5 & 93.8 & $-$ \\
BIER
& 55.3 & 67.2 & 76.9 & 85.1  & $-$ & 78.0 & 85.8 & 91.1 & 95.1 & $-$  \\
\hline
DREML (I,12,48)
& 58.9 & 69.6 & 78.4 & 85.6 & 62.1
& 84.2 & 89.4 & 93.2 & 95.5 & 71.0\\
DREML (R,12,48)
& {\bf63.9} & {\bf75.0} & {\bf83.1} & {\bf89.7} & {\bf67.8}
& {\bf86.0} & {\bf91.7} & {\bf95.0} & {\bf97.2} & {\bf76.4}\\
\hline
\end{tabular}
\end{center}
\label{CUB200CAR196}
\end{table}

\subsection{Retrieval performance}

We follow the evaluation protocol described in~\cite{lifted} to evaluate the Recall@K and Normalized Mutual Information (NMI) values on two datasets, CUB200 and CAR196. For the In-Shop Clothes Retrieval and PKU VehicleID datasets, we follow the evaluation protocol described in~\cite{ICR} and~\cite{VID} and evaluate Recall@K.

\begin{figure}
\begin{center}
{\includegraphics[width=\columnwidth]{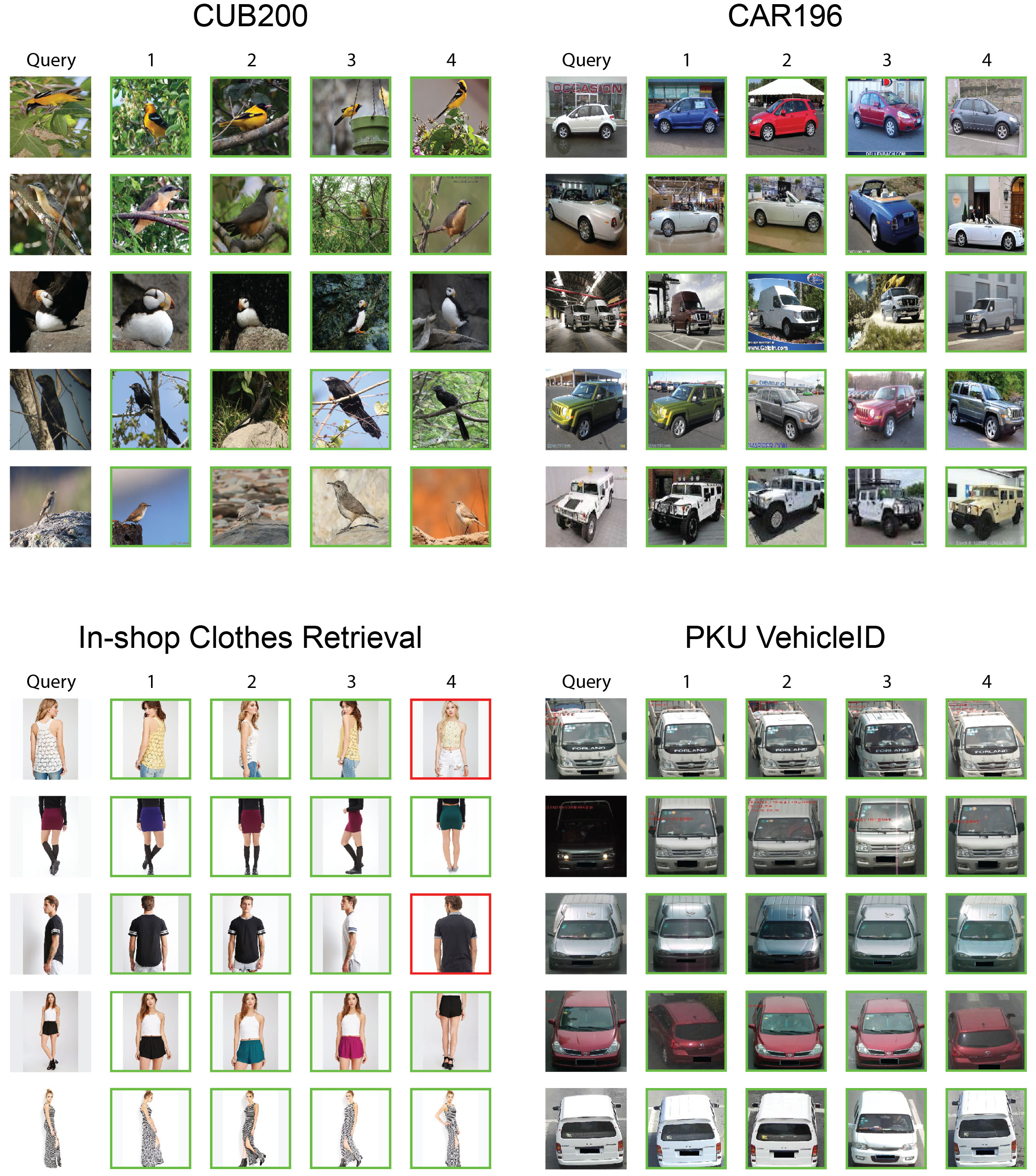}}
\caption{Retrieval results on the CUB200, CAR196, In-Shop Clothes Retrieval and PKU VehicleID dataset. We retrieve the 4 most similar images to the query image. Correct results are highlighted green and incorrect results are highlighted in red.}
\label{exampleImages}
\end{center}
\end{figure}

Table~\ref{CUB200CAR196} shows retrieval performance results on the CUB200 and CAR196 datasets. The {\bf CUB200} dataset~\cite{CUB200} contains 200 classes of birds with 11,788 images. We split the first 100 classes for training (5,864 images) and the rest of the classes for testing (5,924 images). 
The {\bf CAR196} dataset~\cite{CAR196} contains 196 classes of cars with 16,185 images. We use the standard split with the first 98 classes for training (8,054 images) and the rest of the classes for testing (8,131 images). 
For each dataset, Table~\ref{CUB200CAR196} shows the Recall@K for $K=1,2,4,8$. Additionally, the normalized mutual information (NMI) score is included as a measure of clustering performance, as suggested in~\cite{lifted}.  The results show dramatic improvement in both retrieval accuracy and clustering performance across both datasets. Boths datasets contain substantial intra-class variability; in CUB200, birds are shown in different poses in front of very different backgrounds.  Figure~\ref{exampleImages} (top left) highlights this variability and shows example retrieval results from our method for CUB200 (top left) and CAR196 (top right).   

The {\bf In-Shop Clothes Retrieval} (ICR) dataset~\cite{ICR} contains 11,735 classes of clothing items with  54,642 images. Following the settings in~\cite{ICR}, only 7,982 classes of clothing items with 52,712 images are used for training and testing. 3,997 classes are for training (25,882 images) and 3,985 classes are for testing (28,760 images). The test set are partitioned to query set and gallery set, where query set contains 14,218 images of 3,985 classes and gallery set contains 12,612 images of 3,985 classes. Then, given a target image in test set, we retrieve the most similar image in the gallery set.

Table~\ref{ISCresult} shows retrieval and clustering results showing a slight improvement over the BIER results.  In absolute terms, DREML underperforms on the In Shop Clothes dataset compared to other datasets; this dataset has more classes, fewer examples per class, and substantial intra-class variation. Example results showing this variation are shown in Figure~\ref{exampleImages} (bottom left).

\begin{table}
\caption{Retrieval Performance on the In-Shop Clothes dataset.}
\begin{center}
\begin{tabular}{|l|cccc|}
\hline
Method & R@1 & R@10 & R@20 & R@30\\
\hline
FashionNet
& 53.0 & 73.0 & 76.0 & 77.0 \\
HDC
& 62.1 & 84.9 & 89.0 & 91.2 \\
BIER
& 76.9 & 92.8 & 95.2 & 96.2 \\
DREML (R,192,48) & {\bf78.4} & {\bf93.7} & {\bf95.8} & {\bf96.7}\\
\hline
\end{tabular}
\end{center}
\label{ISCresult}
\end{table}

The {\bf PKU VehicleID} (VID)~\cite{VID} dataset contains 221,763 images of 26,267 vehicles captured by surveillance cameras. The training set contains 110,178 images of 13,134 vehicles and the testing set contains 111,585 images of 13,133 vehicles. We follow the standard experimental protocol~\cite{VID} to test on the small, medium and large test set which contains 7,332 images of 800 vehicles, 12,995 images of 1,600 vehicles and 20,038 images of 2,400 vehicles respectively. Table~\ref{VIDresult} shows retrieval and clustering results for the PKU Vehicle-ID dataset.  This dataset has substantially less intra-class variability, but some nearby classes are quite similar.  Example retrieval results and images are shown 
in Figure~\ref{exampleImages} (bottom right).

\begin{table}
\caption{Retrieval Performance on the VID dataset.}
\begin{center}
\begin{tabular}{|l|cc|cc|cc|}
\hline
Data Size & \multicolumn{2}{|c|}{small} & \multicolumn{2}{|c|}{medium} &  \multicolumn{2}{|c|}{large}\\
\hline
Method & R@1 & R@5 & R@1 & R@5 & R@1 & R@5\\
\hline
GS-TRS
& 75.0 & 83.0 & 74.1 & 82.6 & 73.2 & 81.9 \\
BIER
& 82.6 & 90.6 & 79.3 & 88.3 & 76.0 & 86.4 \\
DREML (R,192,12)
& {\bf88.5} & {\bf94.8} & {\bf87.2} & {\bf94.2} & {\bf83.1} & {\bf92.4} \\%
\hline
\end{tabular}
\end{center}
\label{VIDresult}
\end{table}

%%%%%%%%%%%%%%%%%%%%%%%%%%%%%%%%%%%%%%%%%%%%%%%%%%%%%%%%%%%%%%%%%%%%%%%%%%%%%%%%%%%%%%%%%%%%%%%%%%%%%%%%%%%%%%%%%%%%%%%%%%%%%%%%%%%%%%%%%%%%%%%%%%%%%%
\begin{figure}
\begin{center}
{\includegraphics[width=0.6\columnwidth]{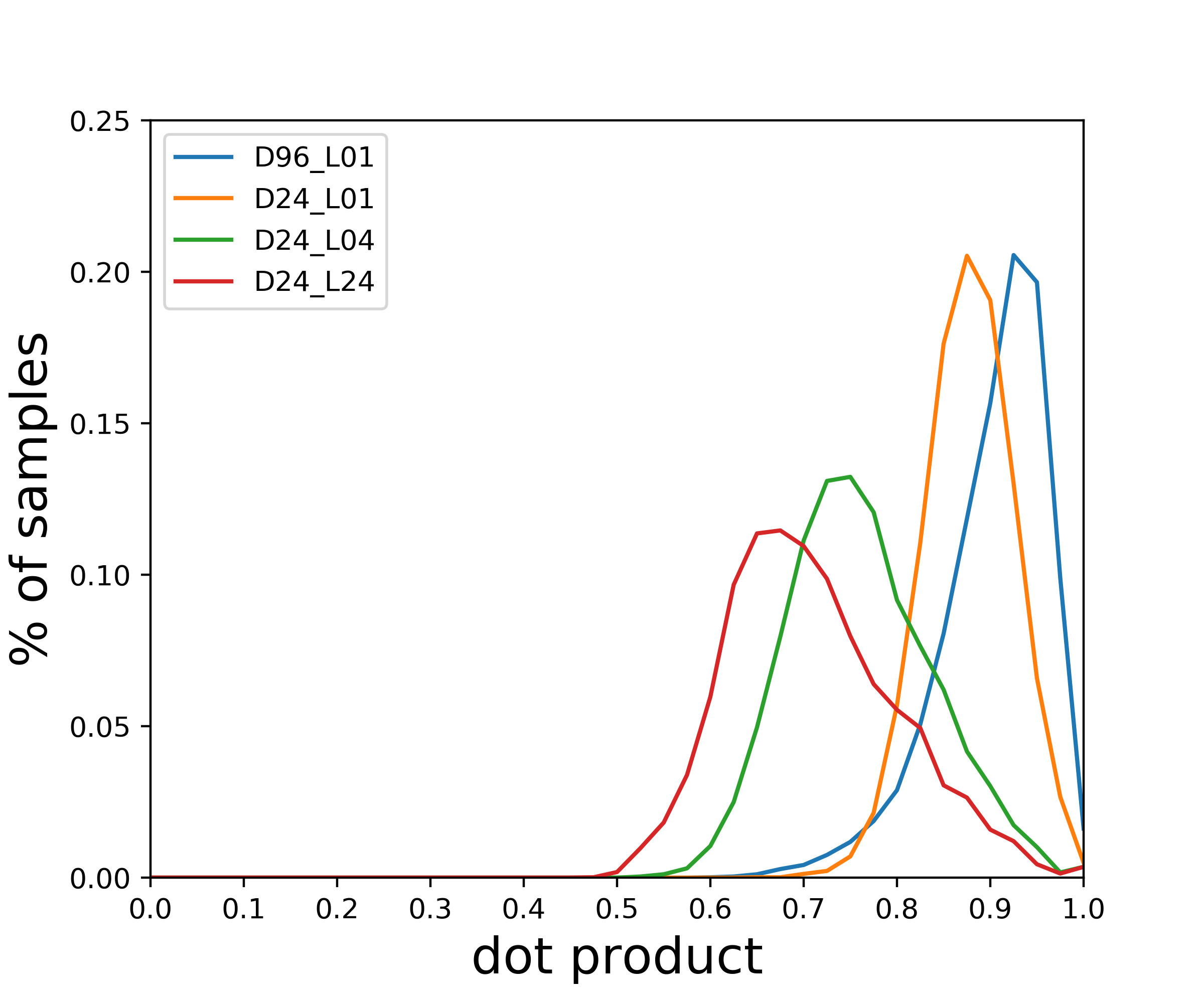}}
\caption{Dot product for each testing image to the closest training image on CAR196 dataset.}
\label{diffusefigure}
\end{center}
\end{figure}

\subsection{Embedding Unseen Classes}
Our approach performs well to embed unseen classes, scattering new examples more effectively across the feature space. We believe that this property helps explain the improved performance of our method on retrieval tasks. We use the No Fuss Embedding approach with 96 training classes to define a 96 dimensional embedding.  We then map inputs from the 98 standard testing classes onto this embedding.   Because the No Fuss Embedding forces points to lie on a hypersphere we use the dot-product as a measure of similarity, and compute the similarity of each point in an unseen class to the most sim ilar point from any of the training classes.

Figure~\ref{diffusefigure} shows this distribution.  The blue line is the distribution for a single network, with most unseen examples having a maximum similarity score to a training class of greater than 0.9.  This shows that when mapping new classes into the embedding space, they are often mapped very close to existing classes, and this crowding within the embedding space may limit recall performance.

We repeat this experiment with 3 other networks.  For the same 96 training categories, we group them into 24 meta-classes, each of size 4 and perform the same experiment (shown in the orange curved, shifted second farthest to the right).  This is not an ensemble embedding, but we hypothesize that the meta-classes comprised of dis-similar input encourages an embedding that pushes novel images farther away from existing images.

The final two curves show the results of the ensemble embedding, using 4 (green) and 24 (red) total embedding functions respectively.  In these embeddings, new images are mapped to locations where they tend to be much farther from the training images, and because they are more spread out the embedding may be more effective at representing unseen categories.

\section{Discussion}
\label{Conclusion}
Deep Randomized Ensembles for Metric Learning (DREML) is a simple approach to creating an ensemble of diverse embedding functions.  We think this is a handy tool that may have broad applicability and have demonstrated results on four datasets spanning problem domains from a medium number of categories in CUB200 and CAR196, to the In-Shop Clothes and Vehicle ID datasets with tens of thousands of categories.  Ensemble based approaches, both ours and BIER paper outperform the non-ensemble approaches by a dramatic margin on all four datasets.

The CARS196 and CUB200 datasets have a moderate amount of training data, and we believe that our approach of building meta-classes creates a version of "label augmentation" that effectively allows our ensemble to have more independent embeddings.  For larger datasets, our approach is similar in performance to BIER for the In-Shop Clothes Retrieval dataset, which has substantial in-class variation due to color and pose changes, and is overall less balanced with many classes that have few examples per class.  We outperform BIER for the PKU Vehicle ID dataset for all dataset sizes, perhaps because our ensemble approach is more robust that BIER to the relatively smaller intra-class variation.  

The downside to our approach is that we train a large number of networks, in the cases where we outperform BIER, we showed results with an ensemble of 12 networks (for In-Shop Clothes dataset) and 48 networks (for the CUB, CAR, and VID dataset), something which affects both the training phase and the test time computational requirements.  It is interesting to explore if the benefits of this ensemble approach can be replicated within a single network.  

\bibliography{references}

\begin{thebibliography}{10}
\providecommand{\url}[1]{\texttt{#1}}
\providecommand{\urlprefix}{URL }
\providecommand{\doi}[1]{https://doi.org/#1}

\bibitem{DBLP:journals/corr/BaiGLWHD17}
Bai, Y., Gao, F., Lou, Y., Wang, S., Huang, T., Duan, L.: Incorporating
  intra-class variance to fine-grained visual recognition. CoRR
  \textbf{abs/1703.00196} (2017), \url{http://arxiv.org/abs/1703.00196}

\bibitem{chen2017beyond}
Chen, W., Chen, X., Zhang, J., Huang, K.: Beyond triplet loss: a deep
  quadruplet network for person re-identification. In: Proc. CVPR. vol.~2
  (2017)

\bibitem{chopra2005learning}
Chopra, S., Hadsell, R., LeCun, Y.: Learning a similarity metric
  discriminatively, with application to face verification. In: Computer Vision
  and Pattern Recognition, 2005. CVPR 2005. IEEE Computer Society Conference
  on. vol.~1, pp. 539--546. IEEE (2005)

\bibitem{guo2015deep}
Guo, J., Gould, S.: Deep cnn ensemble with data augmentation for object
  detection. arXiv preprint arXiv:1506.07224  (2015)

\bibitem{hadsell2006dimensionality}
Hadsell, R., Chopra, S., LeCun, Y.: Dimensionality reduction by learning an
  invariant mapping. In: Computer vision and pattern recognition, 2006 IEEE
  computer society conference on. vol.~2, pp. 1735--1742. IEEE (2006)

\bibitem{resnet}
He, K., Zhang, X., Ren, S., Sun, J.: Deep residual learning for image
  recognition. In: The IEEE Conference on Computer Vision and Pattern
  Recognition (CVPR) (June 2016)

\bibitem{HermansBeyer2017Arxiv}
Hermans*, A., Beyer*, L., Leibe, B.: {In Defense of the Triplet Loss for Person
  Re-Identification}. arXiv preprint arXiv:1703.07737  (2017)

\bibitem{CAR196}
Krause, J., Stark, M., Deng, J., Fei-Fei, L.: 3d object representations for
  fine-grained categorization. In: 4th International IEEE Workshop on 3D
  Representation and Recognition (3dRR-13). Sydney, Australia (2013)

\bibitem{law2013quadruplet}
Law, M.T., Thome, N., Cord, M.: Quadruplet-wise image similarity learning. In:
  Computer Vision (ICCV), 2013 IEEE International Conference on. pp. 249--256.
  IEEE (2013)

\bibitem{VID}
Liu, H., Tian, Y., Wang, Y., Pang, L., Huang, T.: Deep relative distance
  learning: Tell the difference between similar vehicles. In: Proceedings of
  the IEEE Conference on Computer Vision and Pattern Recognition. pp.
  2167--2175 (2016)

\bibitem{Proxy}
Movshovitz-Attias, Y., Toshev, A., Leung, T.K., Ioffe, S., Singh, S.: No fuss
  distance metric learning using proxies. In: The IEEE International Conference
  on Computer Vision (ICCV) (Oct 2017)

\bibitem{lifted}
Oh~Song, H., Xiang, Y., Jegelka, S., Savarese, S.: Deep metric learning via
  lifted structured feature embedding. In: The IEEE Conference on Computer
  Vision and Pattern Recognition (CVPR) (June 2016)

\bibitem{BIER}
Opitz, M., Waltner, G., Possegger, H., Bischof, H.: Bier - boosting independent
  embeddings robustly. In: The IEEE International Conference on Computer Vision
  (ICCV) (Oct 2017)

\bibitem{pytorch}
Paszke, A., Gross, S., Chintala, S., Chanan, G., Yang, E., DeVito, Z., Lin, Z.,
  Desmaison, A., Antiga, L., Lerer, A.: Automatic differentiation in pytorch.
  In: NIPS-W (2017)

\bibitem{ILSVRC15}
Russakovsky, O., Deng, J., Su, H., Krause, J., Satheesh, S., Ma, S., Huang, Z.,
  Karpathy, A., Khosla, A., Bernstein, M., Berg, A.C., Fei-Fei, L.: {ImageNet
  Large Scale Visual Recognition Challenge}. International Journal of Computer
  Vision (IJCV)  \textbf{115}(3),  211--252 (2015).
  \doi{10.1007/s11263-015-0816-y}

\bibitem{schroff2015facenet}
Schroff, F., Kalenichenko, D., Philbin, J.: Facenet: A unified embedding for
  face recognition and clustering. In: Proceedings of the IEEE conference on
  computer vision and pattern recognition. pp. 815--823 (2015)

\bibitem{triplet}
Schroff, F., Kalenichenko, D., Philbin, J.: Facenet: A unified embedding for
  face recognition and clustering. In: The IEEE Conference on Computer Vision
  and Pattern Recognition (CVPR) (June 2015)

\bibitem{Npair}
Sohn, K.: Improved deep metric learning with multi-class n-pair loss objective.
  In: Lee, D.D., Sugiyama, M., Luxburg, U.V., Guyon, I., Garnett, R. (eds.)
  Advances in Neural Information Processing Systems 29, pp. 1857--1865. Curran
  Associates, Inc. (2016)

\bibitem{Szegedy_2016_CVPR}
Szegedy, C., Vanhoucke, V., Ioffe, S., Shlens, J., Wojna, Z.: Rethinking the
  inception architecture for computer vision. In: The IEEE Conference on
  Computer Vision and Pattern Recognition (CVPR) (June 2016)

\bibitem{ustinova2016learning}
Ustinova, E., Lempitsky, V.: Learning deep embeddings with histogram loss. In:
  Advances in Neural Information Processing Systems. pp. 4170--4178 (2016)

\bibitem{CUB200}
Welinder, P., Branson, S., Mita, T., Wah, C., Schroff, F., Belongie, S.,
  Perona, P.: {Caltech-UCSD Birds 200}. Tech. Rep. CNS-TR-2010-001, California
  Institute of Technology (2010)

\bibitem{DBLP:journals/corr/YuanYZ16}
Yuan, Y., Yang, K., Zhang, C.: Hard-aware deeply cascaded embedding. CoRR
  \textbf{abs/1611.05720} (2016), \url{http://arxiv.org/abs/1611.05720}

\bibitem{HDC}
Yuan, Y., Yang, K., Zhang, C.: Hard-aware deeply cascaded embedding. In: The
  IEEE International Conference on Computer Vision (ICCV) (Oct 2017)

\bibitem{ICR}
Ziwei~Liu, Ping~Luo, S.Q.X.W., Tang, X.: Deepfashion: Powering robust clothes
  recognition and retrieval with rich annotations. In: Proceedings of IEEE
  Conference on Computer Vision and Pattern Recognition (CVPR) (June 2016)

\end{thebibliography}
\bibliographystyle{splncs}
\end{document}